\documentclass[10pt,twocolumn,letterpaper]{article}

\usepackage{iccv}      

\usepackage{booktabs,multirow,adjustbox,diagbox,threeparttable,tabularray}
\definecolor{iccvblue}{rgb}{0.21,0.49,0.74}
\usepackage[pagebackref,breaklinks,colorlinks,allcolors=iccvblue]{hyperref}
\usepackage{colortbl}
\usepackage[accsupp]{axessibility}
\def\paperID{11584} 
\def\confName{ICCV}
\def\confYear{2025}

\title{DynImg: Key Frames with Visual Prompts are Good Representation for Multi-Modal Video Understanding}

\author{%
  Xiaoyi Bao$^{1,2,3}$, Chenwei Xie$^{3}$, Hao Tang$^{3,4}$, Tingyu Weng$^{3}$,\\
  Xiaofeng Wang$^{1,2}$,
  Yun Zheng$^{3}$, Xingang Wang$^{1,5}$\thanks{Corresponding author.}\\ 
  $^1$Institute of Automation, Chinese Academy of Sciences \\
  $^2$School of Artificial Intelligence, University of Chinese Academy of Sciences \\ 
  $^3$Alibaba Group \quad $^4$Peking University  \quad $^5$Luoyang Institute for Robot and Intelligent Equipment\\
  Project page: \href{https://dynimg.github.io/}{https://dynimg.github.io/}
}

\begin{document}

\twocolumn[{%
\renewcommand\twocolumn[1][]{#1}%
\maketitle
\vspace{-15pt}
\centering
\includegraphics[width=0.98\linewidth]{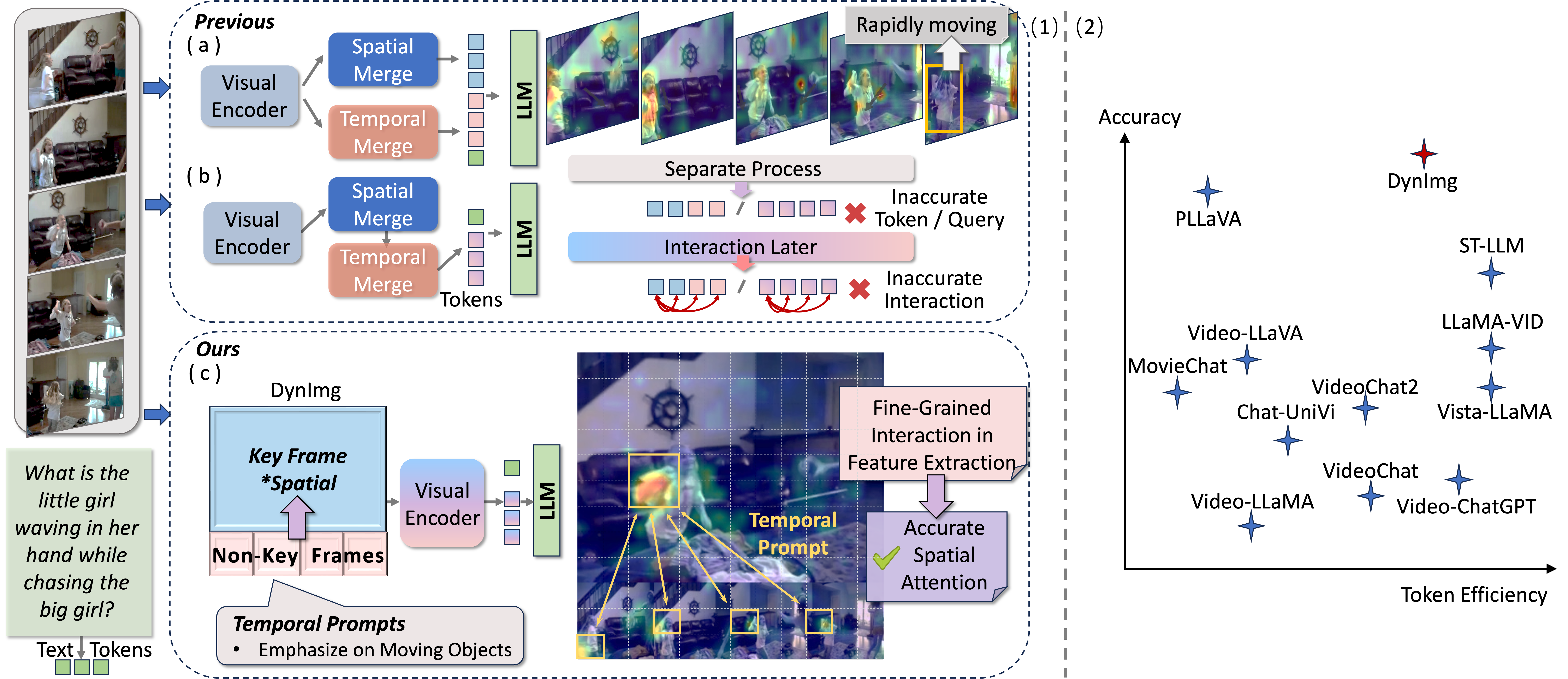}
\captionof{figure}{Comparison between the previous methods for video understanding and DynImg. As for the structure comparison in the left (1), previous models use processed visual features for subsequent spatial and temporal merging modules, either in parallel (a) or in sequence (b). However, in rapidly moving scenarios like the example on the right, where the little girl quickly turns back and moves in the last frame, these models fail to capture the crucial details of her motion. Factors such as motion blur lead to these important temporal details being overlooked during the visual feature extraction process, resulting in these areas not receiving the necessary attention. Spatio-temporal interaction based on such inaccurate features or tokens is ineffective. In contrast, our proposed DynImg (c) advances the spatio-temporal interaction process via temporal prompts. This enables the model to focus on those rapidly moving regions that are difficult to capture during the feature extraction phase. The right part shows its effectiveness and efficiency (2). ``Accuracy'' is the average accuracy on MSVD, MSRVTT, Activitynet, and TGIF. ``Token Efficiency'' is negatively correlated with the number of visual tokens used to represent the video.}
\label{first}
\vspace{18pt}
}]

\begin{abstract}
In recent years, the introduction of Multi-modal Large Language Models (MLLMs) into video understanding tasks has become increasingly prevalent.
However, how to effectively integrate temporal information remains a critical research focus.
Traditional approaches treat spatial and temporal information separately. Due to issues like motion blur, it is challenging to accurately represent the spatial information of rapidly moving objects. This can lead to temporally important regions being underemphasized during spatial feature extraction, which in turn hinders accurate spatio-temporal interaction and video understanding.
To address this limitation, we propose an innovative video representation method called Dynamic-Image (DynImg).
Specifically, we introduce a set of non-key frames as temporal prompts to highlight the spatial areas containing fast-moving objects.
During the process of visual feature extraction, these prompts guide the model to pay additional attention to the fine-grained spatial features corresponding to these regions.
Moreover, to maintain the correct sequence for DynImg, we employ a corresponding 4D video Rotary Position Embedding. This retains both the temporal and spatial adjacency of DynImg, helping MLLM understand the spatio-temporal order within this combined format.
Experimental evaluations reveal that DynImg surpasses the state-of-the-art methods by approximately 2\% across multiple video understanding benchmarks, proving the effectiveness of our temporal prompts in enhancing video comprehension. https://dynimg.github.io/
\vspace{-10pt}
\end{abstract}   
\section{Introduction}
\label{sec:intro}
With the widespread application of Multi-modal Large Language Models (MLLMs) in image-based vision-language tasks~\cite{llava,alayrac2022flamingo,li2023blip,zhu2023minigpt,lai2023lisa,bao2024cores,tang2025ufo}, there has been a growing research interest in extending their application to video understanding~\cite{lin2023videollava,zhang2023videollama,li2023videochat,liu2025hybrid}. However, when dealing with videos instead of static images, the complexity increases significantly due to the need to capture temporal information.

Previous methods typically decouple the processing of temporal and spatial information \cite{maaz2023videochatgpt,zhang2023videollama,jin2024chatuniv,li2023videochat}. Their spatio-temporal interaction occurs within high-level queries or tokens. This corresponds to the two different model structures in Fig.\ref{first}(a/b), both of them using a pre-trained image encoder to extract spatial features in the beginning.
In Fig.\ref{first}(a), after compression through temporal and spatial modules, temporal tokens and spatial tokens interact within the LLM in parallel. For instance, the two compression in Video-ChatGPT \cite{maaz2023videochatgpt} refer to different spatial and temporal pooling modules.
The other struction, as is shown in Fig.\ref{first}(b), processes temporal features sequentially. Here, spatio-temporal interaction is based on the extracted high-level spatial features. An example is Video-LLaMA \cite{zhang2023videollama}, which extracts frame-level embeddings using Q-former and then utilizes an additional temporal module for inter-frame temporal interaction.

However, performing spatio-temporal interaction within high-level queries or tokens may result in suboptimal performance due to the loss of fine-grained information. Throughout visual feature extraction and spatial merging, spatial information undergoes continuous abstraction and compression. Techniques such as k-means clustering \cite{jin2024chatuniv}, pooling \cite{li2025llamavid,maaz2023videochatgpt}, Q-former \cite{zhang2023videollama,li2023videochat}, or simple convolution \cite{lin2023videollava} cause many fine-grained details to blur and be lost during averaging.
This issue is particularly noticeable in regions containing rapidly changing objects. Due to factors like motion blur, such areas often fail to gain an accurate and comprehensive fine-grained feature representation. As illustrated in the upper right of Fig.\ref{first}(1), the attention on the area where a young girl suddenly turns back and moves in the last frame is noticeably lacking. Such rapidly changing regions are temporally significant for video understanding. If their fine-grained features are neglected in the initial spatial extraction, the effectiveness of subsequent interactions, performing at the query or token level, becomes severely limited. This diminishes video comprehension in complex or dynamically moving scenes.

To address this challenge, we need to implement explicit spatio-temporal interactions from the initial stage of the visual feature extraction.
We hope that the visual encoder can identify which regions or objects are rapidly moving from the video frames.
In multi-modal image understanding, visual prompts ( \eg, bounding boxes and sequence labels) are commonly employed to convey object-centric textual information within images~\cite{prompt45,prompt11,promptfirst,prompt31,prompt8,prompt32}. These prompts augment input images with supplementary information and adjust the focus to the text-referenced objects being prompted.
Drawing inspiration from this, we propose our DynImg. We propose to construct a temporal prompt to emphasize regions containing rapidly moving objects during the feature extraction phase.
Specifically, we decompose video sequences into keyframes and non-keyframes, relying on the keyframe content for high-resolution spatial representation. The crucial temporal prompt component is constructed by overlaying certain non-keyframes onto the original images. The effectiveness of the prompts occurs within the encoder, leveraging the long-range modeling capability inherent in the visual encoder~\cite{dosovitskiy2020image,clip,siglip}. These prompts adjust the attention on the keyframe to ensure that regions containing rapidly moving objects are properly emphasized rather than overlooked. In this way, it facilitates effective fine-grained spatiotemporal interaction.

With the spatial and temporal correspondences in prompts and keyframes,
DynImg enriches the original spatial domain with an added temporal dimension during the feature extraction. However, for LLM, the form of DynImg is so unfamiliar that directly use the obtained final token of DynImg may introduce spatiotemporal disarray. To avert this sequential confusion, we propose a 4D positional embedding mechanism. This mechanism constructs a unified coordinate system for the input sequence, especially for the visual tokens corresponding to DynImg.
Unlike traditional 1D-RoPE~\cite{llava} that focuses only on token sequence, our proposed 4D-RoPE maintains the correct four-dimensional sequence of DynImg. It not only ensures the consistency in adjacency relations within both the spatial and temporal domains, but also guides MLLM to understand the composition logic of the DynImg format.

Our contributions can be summarized as follows:
\begin{itemize}
\item We introduce Dynamic-Image (DynImg), a novel video representation method. DynImg addresses the shortcomings of traditional approaches by facilitating fine-grained spatiotemporal attention interactions. It ensures that regions containing rapidly moving objects are appropriately emphasized during the initial visual feature extraction, thereby improving overall video comprehension.

\item We propose using non-key frames as temporal prompts. These prompts inform the visual encoder about regions containing rapidly moving objects, allowing it to pay extra attention to these areas in fine-grained feature extraction of keyframes. This is further supported by our proposed novel 4D Video Rotary Position Embedding. It guides the temporal and spatial sequence of DynImg, aiding the LLM in accurately understanding DynImg's composition.

\item Extensive experiments demonstrate the effectiveness and efficiency of our proposed method, achieving state-of-the-art accuracy on most Video-QA benchmarks.
\end{itemize}
\section{Related Work}
\label{sec:formatting}

\subsection{Video Temporal Representation}

\begin{figure*}[t]
\centering
\includegraphics[width=0.95\linewidth]{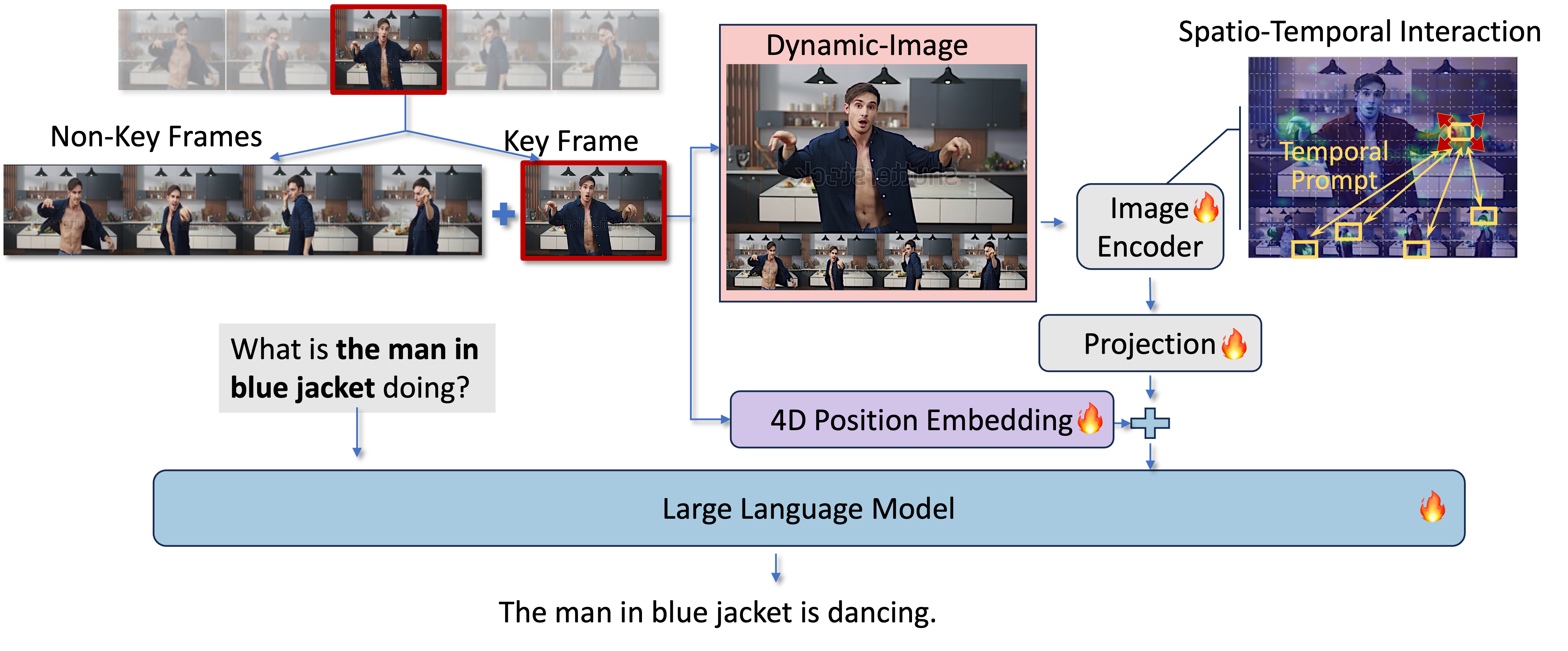}
\vspace{-5pt}
\caption{Overall architecture of DynImg. 
Videos are decomposed into keyframes and non-keyframes. Several non-keyframes serve as temporal prompts and are combined with keyframes to form the representation of the DynImg, along with its corresponding 4D positional embeddings. Within the image encoder, these temporal prompts adjust the spatial attention. The attention map on the left belongs to the patch of the left hand in the keyframe. Red arrows indicate the usual influence of local visual features, while yellow arrows show the emphasis to dynamic regions by non-keyframes. The output feature of the image encoder, after passing through a projection layer, is fed into the LLM along with the positional embeddings for the final output.
}
\label{overall}
\end{figure*}

Since DeepVideo~\cite{deepvideo} first applied CNN networks to video understanding, researchers have been focusing on how to effectively extract and utilize the sequential representation of videos. One approach involves incorporating traditional handcrafted features to represent sequences. For example, Simonyan introduces a Two-Stream structure where optical flow is employed, and object motion features are learned through an additional temporal stream network~\cite{twostream}. Following this, subsequent studies enhance the performance of dual-stream networks. For instance, LSTM is used for extended temporal modeling~\cite{twolstm}, and early fusion techniques and 3D pooling are applied to bolster the fusion of sequential information~\cite{twoearly}. TDD overlays optical flow features along trajectories to achieve superior results~\cite{tdd}. In addressing long video inputs, TSN divides input into segments, processing them through a dual-stream network before merging~\cite{tsn}. Research efforts such as DOVF~\cite{dovf} and TLE~\cite{tle} extend the functional scope of dual-stream networks by proposing and refining global encodings. Additionally, Yang suggests substituting optical flow with motion vectors~\cite{motion1}, while Video-LaVIT~\cite{jin2024videolavit} further optimizes the use of action vectors.

Another category of methods includes C3D~\cite{c3d} and I3D~\cite{i3d}, which employ 3D convolution to model the entire spatiotemporal structure of videos. R(2+1)D~\cite{rtwooned} decomposes 3D CNNs into 2D spatial convolutions and 1D temporal convolutions, reducing overfitting and training difficulty. SlowFast~\cite{slowfast} uses a combination of fast and slow networks for video classification, with each network learning static and motion information, respectively. Methods like TimeSformer~\cite{timesformer} and ViViT~\cite{vivit} split the joint attention into temporal and spatial parts to reduce memory usage.

\subsection{Video Understanding}
Video understanding typically manifests as tasks such as video question answering and video captioning, involving visual text interactions. These tasks require accurate comprehension of video content based on textual and video inputs, yielding pure text responses. Recently, in response to higher-dimensional visual understanding demands, researchers have begun leveraging visual knowledge contained within multimodal large language models to aid in the abstraction of visual information.

A series of research efforts initially decompose videos into different representation dimensions and then integrate the inputs to enrich the prompts for MLLM. For example, Video-ChatGPT~\cite{maaz2023videochatgpt} divides videos into spatial and temporal branches for pooling. VideoChat~\cite{li2023videochat} breaks down videos into textual descriptions and feature embeddings of the videos themselves. LLaMA-VID~\cite{li2025llamavid} represents each frame as two tokens: context markers and content markers. Video-LaViT~\cite{jin2024videolavit} employs keyframes and motion vectors to tokenize videos. PLLaVA~\cite{xu2024pllava} and PPLLaVA~\cite{liu2024ppllava} propose adaptive pooling and Prompt-guided Pooling to reduce spatiotemporal redundancy. IG-VLM~\cite{kim2024IGVLM} uses a similar comic-style image grid, leveraging GPT-4V to achieve improved performance. Simultaneously, to augment training data, researchers have explored methods to unify image modalities for training. Chat-UniVi~\cite{jin2024chatuniv} and Video-LLaVA~\cite{lin2023videollava}, respectively, achieve this by employing object-based adaptive clustering tokens and alignment followed by projection strategies, thereby unifying image and video inputs for enhanced visual understanding.

\subsection{Visual Prompt}

Prompting has been extensively studied in the natural language processing community~\cite{promptnlp}. Recently, researchers have begun to explore the benefits of using prompting in image recognition. This involves adding learnable modifications to the image to guide the model to make specific predictions~\cite{prompt45,prompt11,promptfirst,prompt31,prompt8}. These prompt-tuning methods optimize an additional and fixed visual prompt, such as adding an optimizable pixel region around the image~\cite{prompt42,prompt44,khattak2023maple,bao2024relevant}. Recently, researchers have shown that manually crafted prompts, such as red circles and blurred backgrounds, can effectively guide the attention of models like CLIP~\cite{clip} when training on datasets containing similar annotations~\cite{prompt32}. Additionally, there is also research suggesting that constructing embeddings corresponding to the text prompts and appending them to the original image can better facilitate information supplementation and alignment. However, the aforementioned studies have mainly focused on static image-related tasks and have not been extended to video inputs.

\section{Methods}
To mitigate the information loss associated with coarse-grained interactions, we propose our Dynamic-Image (DynImg) model. Rather than abstracting features separately along spatial and temporal dimensions, we prioritize the spatiotemporal interaction process. Inspired by the information exchange in visual prompts, we introduce temporal prompts for this fine-grained level of spatiotemporal interaction. In Section 3.1, we describe the composition of DynImg and the functioning process of the temporal prompts. Since the prompts introduce a temporal dimension within the spatial framework, we propose a corresponding 4D rotary positional encoding to maintain correct temporal and spatial order. In Section 3.2, we discuss the implementation and details of this positional embedding mechanism.

\subsection{Temporal Prompts}
If spatial information is extracted and integrated before performing spatiotemporal interactions, many crucial spatial details may have already been discarded or blurred before the interaction. This includes temporally significant local regions, such as foreground objects undergoing displacement.
How can these temporally important fine-grained details be retained? We propose to advance the spatiotemporal interaction process, changing the interaction targets from abstract tokens or features to fine-grained pixel values.

How, then, can temporal information be integrated at the pixel level for interaction? In multimodal image tasks, visual prompts are utilized for text-related target emphasis. This often involves adding bounding boxes, labels, or other annotations directly onto the original image. Inspired by this idea, we propose a video-specific temporal prompt that integrates temporal information to adjust attention to spatial details. For moving targets within the temporal prompt, we identify and emphasize the corresponding spatial regions, enabling fine-grained spatiotemporal interactions for enhanced understanding of spatial content.

\noindent\textbf{Composition of the Dynamic-Image}: The structure of DynImg involves decoupling the video into keyframes and non-keyframes, relying on keyframes to provide a fine-grained spatial basis while utilizing a set of non-keyframes as temporal prompts. 

For keyframe selection, we apply the MPEG-4 method wherein I-frames inherently contain more information. We evenly sample four I-frames as keyframes $K$. 
For a given key-frame $K_i$, we randomly select two preceding I-frames from $(K_{i-1},K_{i})$ and two following it from $(K_{i},K_{i+1})$ as non-keyframes ($N$-frames), forming a frame-group to establish a comprehensive temporal context for $K_i$. Overlap may happen in non-keyframe selection of two adjacent DynImgs, while keyframes appear only once.

To preserve fine-grained spatial details, high-resolution keyframes are retained as the base graph for DynImg. For the prompt portion, we resize the four $N$-frames and concatenate them in temporal order from left to right beneath the $K$ frame after applying data augmentation to each whole group. As the temporal prompt mainly focuses on motion, the resizing operation does not significantly impact its motion-aware effectiveness. This resizing is controlled to ensure patch-based feature extraction does not cross the boundaries between frames after concatenation.

\noindent\textbf{Functioning of temporal prompts}: The DynImg, carrying both temporal and spatial information, is fed into the video feature extraction pipeline. 
In addition to focusing on the spatial features of adjacent local patches, the visual encoder utilizes long-range modeling capabilities to facilitate interaction between the spatially informative keyframe regions and the prompts positioned below them. 

Specifically, within the self-attention layer of the vision transformer in the visual encoder, patches from keyframes can attend to highly similar patches within the temporal prompt regions. 
For targets with temporal variations, their corresponding patches from the $N$-frames undergo deformations and movements across different frames. The keyframe patches identify movement trends from the prompts. They establish connections, adding temporal dynamics to fine-grained spatial features. 
Through task-driven training, the model progressively increases attention weights on dynamic local regions. This makes the dynamic changes in temporal prompts effectively activate the fine-grained regions in keyframes.

\begin{figure}
\centering
\includegraphics[width=0.7\linewidth]{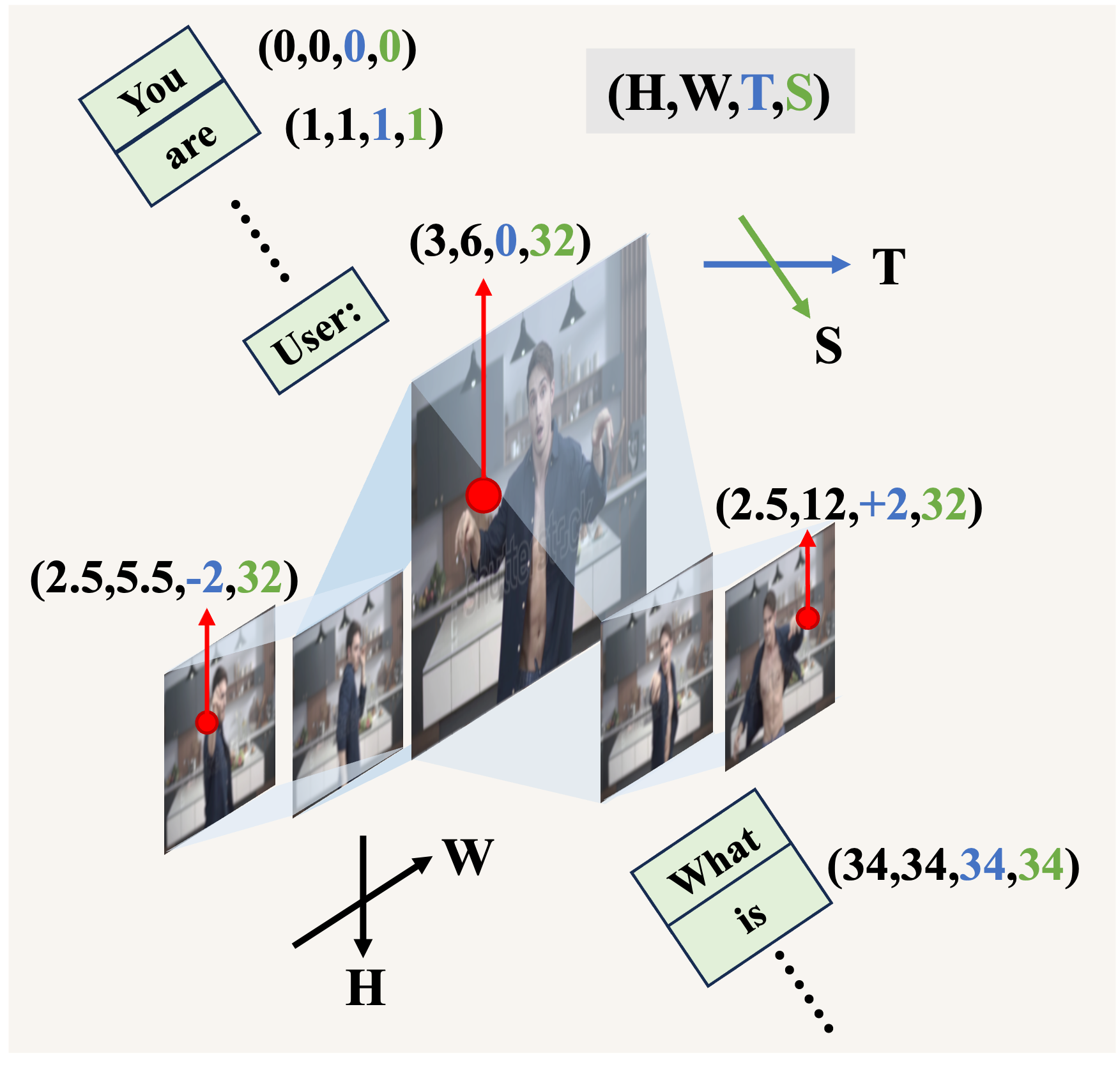}
\vspace{-5pt}
\caption{Schematic diagram of coordinates in 4D position embedding. The direction of the arrow represents the direction of increase in the four dimensions of height, weight, temporal, and sequence. The first two dimensions of the visual prompt are maintained through interpolation to preserve spatial correlation. The temporal dimension has a coordinate value of 0 at the keyframe time point and undergoes symmetrical changes before and after in the visual prompt section. The same dynamic image corresponds to the same sequence coordinate.}
\label{rope}
\vspace{-5pt}
\end{figure}

\subsection{4-Dimensional Position Embedding}

\begin{table*}[!t]
\centering
\setlength{\tabcolsep}{8pt}
\begin{threeparttable}
\resizebox{\textwidth}{!}{
\begin{tabular}{l|c|c|cc|cc|cc|cc|cccccc}
\toprule[1pt]
\multirow{2}{*}{Method}&\multirow{2}{*}{Encoder}&\multirow{2}{*}{LLM}&\multicolumn{2}{c|}{MSVD}&\multicolumn{2}{c|}{MSRVTT}&\multicolumn{2}{c|}{Activitynet}&\multicolumn{2}{c|}{TGIF}&\multicolumn{6}{c}{Video-ChatGPT}\\
&&&Acc&Score&Acc&Score&Acc&Score&Acc&Score&CI&DO&CU&TU&CO&Avg.\\
\midrule
Video-LLaMA~\cite{zhang2023videollama}&BLIP2&7B&51.6&2.5&29.6&1.8&12.4&1.1&-&-&1.96&2.18&2.16&1.82&1.79&1.98\\
Video-ChatGPT~\cite{maaz2023videochatgpt}&ViT-L&7B&64.9&3.3&49.3&2.8&35.2&2.7&51.4&3.0&2.50&2.57&2.69&2.16&2.20&2.42\\
Video-LLaVA~\cite{lin2023videollava}&ViT-L&7B&70.7&3.9&59.2&3.5&45.3&3.3&70.0&4.0\\
Chat-UniVi~\cite{jin2024chatuniv}&CLIP-L&7B&65.0&3.6&54.6&3.1&45.8&3.2&60.3&3.4&2.89&2.91&3.46&2.89&2.81&2.99\\
MovieChat~\cite{song2024moviechat}&BLIP2&7B&75.2&3.8&52.7&2.6&45.7&3.4&-&-&2.76&2.93&3.01&2.24&2.42&2.67\\
VideoChat~\cite{li2023videochat}&BLIP2&7B&56.3&2.8&45.0&2.5&26.5&2.2&34.4&2.3&2.23&2.50&2.53&1.94&2.24&2.29\\
VideoChat2~\cite{li2024mvbench}&UMT-L&7B&70.0&3.9&54.1&3.3&49.1&3.3&-&-&3.02&2.88&3.51&2.66&2.81&2.98\\
Vista-LLaMA~\cite{ma2023vista}&BLIP2&7B&65.3&3.6&60.5&3.3&48.3&3.3&-&-&2.44&2.64&3.18&2.26&2.31&2.57\\
LLaMA-VID~\cite{li2025llamavid}&BLIP2&13B&70.0&3.7&58.9&3.3&47.5&3.3&-&-&2.96&3.00&3.53&2.46&2.51&2.89\\
LITA~\cite{huang2024lita}&CLIP-L&7B&-&-&-&-&-&-&-&-&2.94&2.98&3.43&2.68&\textbf{3.19}&3.04\\
ST-LLM~\cite{liu2024stllm}&BLIP2&7B&74.6&3.9&63.2&3.4&50.9&3.3&-&-&3.23&\textbf{3.05}&\textbf{3.74}&2.93&2.81&3.15\\
IG-VLM~\cite{kim2024IGVLM}&Unk&GPT-4V&76.3&4.0&63.8&3.5&57.0&3.5&65.3&3.7&\textbf{3.40}&2.80&3.61&2.89&3.13&3.17\\
PLLaVA~\cite{xu2024pllava}&ViT-L&7B&76.6&4.1&62.0&3.5&56.3&3.5&77.5&\textbf{4.1}&3.21&2.86&3.62&2.33&2.93&3.12\\
\midrule
\rowcolor{gray!20}
DynImg&SigLip&7B&\textbf{78.6}&\textbf{4.2}&\textbf{64.1}&\textbf{3.5}&\textbf{57.9}&\textbf{3.6}&\textbf{77.5}&4.0&3.33&3.02&\textbf{3.75}&\textbf{2.96}&3.17&\textbf{3.25} \\
\bottomrule[1pt]     
\end{tabular}
}
\end{threeparttable}
\caption{Performance comparison between DynImg and other methods on five open-ended video understanding benchmarks. ``Encoder'' refers to the type of visual encoder used during training.}
\label{mainexcel}
\end{table*}

In the composition of DynImg, different frames correspond to various positions along the temporal dimension. An explicit temporal sequence aids in establishing cross-frame associations between spatially adjacent features. This is particularly useful for identifying spatial correspondences between the temporal prompts and keyframes. Additionally, at the token level, the temporal order, especially among the temporal prompts, plays a crucial role in video understanding. To address these challenges, we propose a 4D rotational positional embedding mechanism tailored for DynImg. This mechanism prevents the confusion of spatio-temporal relationships due to DynImg composition, thereby facilitating MLLM's understanding of this input format.

Rotary Position Embedding(RoPE) is a common positional encoding method used in LLMs. It involves multiplying features $q$ with trigonometric functions based on their positional coordinates $x$. In the original 1D RoPE, features $q$ are split into pairs $(q_1,q_2)$. The multiplication by the rotation matrix is represented by the following formula:
\begin{equation}
R = \begin{pmatrix} q_1 \\ q_2 \end{pmatrix} \otimes \begin{pmatrix} \cos(x\theta) \\ \cos(x\theta) \end{pmatrix} + \begin{pmatrix} -q_2 \\ q_1 \end{pmatrix} \otimes \begin{pmatrix} \sin(x\theta) \\ \sin(x\theta) \end{pmatrix}
\label{eqation}
\end{equation}

where \( \theta \) is a parameter that controls the rotational period, calculated by a formula. \( x \cdot \theta \) gives the rotation angle.

For DynImg, it is necessary to maintain the position order in four dimensions: Height, Weight, Temporal, and Sequence, as shown in Fig.~\ref{rope}. For the text portion of the input token sequence, the sequence coordinate increments from 0, while the coordinates of the first three dimensions $(H,W,T,S)$ are simply set to identical to the fourth.

As for the $(H,W)$ coordinates of visual tokens, the coordinates for the keyframe region remain consistent with those of static images, incrementing spatially from 0. For the visual prompt, to preserve its spatial adjacency across different frames, interpolation is applied to the keyframe's coordinates to obtain coordinates of the first two dimensions. All frames in one DynImg share the same spatial coordinate range. As for the temporal dimension coordinate $(T)$, we set the base keyframe's coordinate as 0, while the coordinate of temporal prompts increase or decrease symmetrically according to their chronological order. All frames of the same dynamic image share the same sequence coordinate $(S)$ that increases as text tokens.

To leverage the pre-trained LLMs using RoPE while adapting to the special format of DynImg, we calculate the angle of rotation via a weighted sum of four-dimensional coordinates.
\begin{equation}
x \cdot \theta = x_h \cdot \theta_h + x_w \cdot \theta_w + x_t \cdot \theta_t + x_s \cdot \theta_s
\label{eqation2}
\end{equation}

where \( x_h \), \( x_w \), \( x_t \), and \( x_s \) represent the coordinates of $(H,W,T,S)$, respectively. For $\theta_s$, we retain the general sinusoidal encoding scheme used by LLMs, calculated as $\theta^i_s=10000^{-2i/d}$. During training, we set \( \theta_h \), \( \theta_w \), and \( \theta_t \) as learnable parameters with initial values of 0. Initially, only \( m_s \cdot \theta_s \) takes effect, ensuring that the model does not disrupt the effectiveness of the pre-trained LLM in the early training phase. Through training, the model gradually learns suitable embedding for \( \theta_h \), \( \theta_w \), and \( \theta_t \).

\section{Experiments}

\subsection{Implementation Details}
\textbf{Training Data}: Following ~\cite{xu2024pllava}, we utilize a series of video-text data pairs for training. The dataset consists of the following categories: Video caption data comprises 39k samples from TextVR, 8k samples from YouCook2, 7k samples from VideoChat~\cite{li2023videochat}, and 400k samples from WebVid. Video classification data includes 40k samples from Kinetics-710 and 40k samples from Something-Something v2. Video conversation data consists of 9k samples from VideoChat2~\cite{li2024mvbench}, 13k samples from Video-ChatGPT~\cite{maaz2023videochatgpt}, and 4k samples from VideoChat. Video reasoning data originates from 34k samples in Next-QA, and 43k samples in CLEVRER-QA. Video question-and-answer data comes from 8k samples in Ego-QA, and 92k samples in TGIF-QA. 
\begin{table*}[!t]
\centering
\begin{threeparttable}
\resizebox{\textwidth}{!}{
\begin{tabular}{l|c|c|ccccccccccccccccccccc}
\toprule[1pt]
\multirow{2}{*}{Method}&Visual&LLM&\multirow{2}{*}{AS}&\multirow{2}{*}{AP}&\multirow{2}{*}{AA}&\multirow{2}{*}{FA}&\multirow{2}{*}{UA}&\multirow{2}{*}{OE}&\multirow{2}{*}{OI}&\multirow{2}{*}{OS}&\multirow{2}{*}{MD}&\multirow{2}{*}{AL}&\multirow{2}{*}{ST}&\multirow{2}{*}{AC}&\multirow{2}{*}{MC}&\multirow{2}{*}{MA}&\multirow{2}{*}{SC}&\multirow{2}{*}{FP}&\multirow{2}{*}{CO}&\multirow{2}{*}{EN}&\multirow{2}{*}{ER}&\multirow{2}{*}{CI}&\multirow{2}{*}{Avg.}\\
 &Tokens&Size&&&&&&&&&&&&&&&&&&&&& \\
\midrule

Video-LLaMA~\cite{lin2023videollava}&2048&7B&27.5&25.5&51.0&29.0&39.0&48.0&40.5&38.0&22.5&22.5&43.0&34.0&22.5&32.5&45.5&32.5&40.0&30.0&21.0&37.0&34.1\\
Video-ChatGPT~\cite{maaz2023videochatgpt}&356&7B&23.5&26.0&62.0&22.5&26.5&54.0&28.0&40.0&23.0&20.0&31.0&30.5&25.5&39.5&\textbf{48.5}&29.0&33.0&29.5&26.0&35.5&32.7\\
VideoChat~\cite{li2023videochat}&1536&7B&33.5&26.5&56.0&33.5&40.5&53.0&40.5&30.0&25.5&27.0&48.5&35.0&20.5&42.5&46.0&26.5&41.0&23.5&23.5&36.0&35.5\\
VideoChat2~\cite{li2024mvbench}&1536&7B&66.0&47.5&83.5&\textbf{49.5}&60.0&58.0&71.5&\textbf{42.5}&23.0&23.0&\textbf{88.5}&39.0&42.0&58.5&44.0&\textbf{49.0}&36.5&35.0&40.5&\textbf{65.5}&51.1\\
ST-LLM~\cite{liu2024stllm}&256&7B&66.0&53.5&\textbf{84.0}&44.0&58.5&80.5&73.5&38.5&42.5&31.0&86.5&36.5&56.5&78.5&43.0&44.5&46.5&34.5&41.5&58.5&54.9\\
IG-VLM~\cite{kim2024IGVLM}&Unk&GPT-4&55.5&\textbf{63.5}&72.0&46.5&\textbf{73.5}&18.5&59.0&29.5&12.0&\textbf{40.5}&83.5&39.0&12.0&22.5&45.0&47.5&52.0&31.0&\textbf{59.0}&11.0&43.5\\
PLLaVA~\cite{xu2024pllava}&2304&7B&58.0&49.0&55.5&41.0&61.0&56.0&61.0&36.0&23.5&26.0&82.0&39.5&42.0&52.0&45.0&42.0&\textbf{53.5}&30.5&48.0&31.0&46.6\\
\midrule
\rowcolor{gray!20}
DynImg&576&7B&\textbf{66.5}&54.0&80.5&49.0&58.5&\textbf{82.0}&\textbf{74.0}&38.0&\textbf{44.5}&33.5&79.5&\textbf{41.0}&\textbf{57.0}&\textbf{78.5}&43.5&46.0&42.5&\textbf{39.0}&51.0&57.0&\textbf{55.8} \\
\bottomrule[1pt]
\end{tabular}
}
\end{threeparttable}
\caption{Experiments on MVBench, the multi-choice question answering dataset. The 20 types of questions included require an overall understanding of the video content for their answers.}
\label{mvbench}
\end{table*}

\noindent\textbf{Evaluation Setting}: For the evaluation dataset, we follow the previous methods and use two kinds of video understanding tasks~\cite{lin2023videollava,xu2024pllava}. For the open-ended video question-and-answer dataset, we select MSVD~\cite{msvd}, MSRVTT~\cite{msrvtt}, TGIF~\cite{li2016tgif}, and ActivityNet~\cite{activitynet}. The correctness and confidence scores of generated answers corresponds to the ``Score'' in Tab.~\ref{mainexcel}, a float value in (0,5) output directly by 
GPT. Additionally, we use the benchmark from Video-ChatGPT~\cite{maaz2023videochatgpt}, which is also a GPT-assisted evaluation in five dimensions: CI (Correctness of Information), DO (Detail Orientation), CU (Context Understanding), TU (Temporal Understanding), and CO (Consistency). The GPT model we use for evaluation is GPT-3.5-turbo-0125. As for the performance on the multi-choice question-and-answer task, we choose the dataset MVBench~\cite{li2024mvbench}. This dataset evaluates video understanding performance across 20 different domain, where each domain contains 200 multiple-choice instances. For example, the ``AS'' in Tab.~\ref{mvbench} refers to Action Sequence. The accuracy calculation does not rely on GPT.

\noindent\textbf{Models Setting}: We choose four $N$-frame as temporal prompt for one keyframe in each DynImg. We first perform regular data augmentation (RandomResizedCrop + RandomHorizontalFlip + normalization) on these five frames together before composing them to be one DynImg. This step aims to prevent the spatial region confusion caused by the central crop operation during the data augmentation process. Considering the video length, we synthesize four DynImgs as input. We use a pre-trained Siglip-so400m-384~\cite{siglip} as our video encoder. As for the projection layer, it consists of a feedforward layer and the Adaptive Average Structure Pooling module from PLLaVA~\cite{xu2024pllava}, with the pooling shape set to (16, 12, 12). The training recipe of DynImg follows PLLaVA. For the LLM selection, we use the pretrained Qwen2.5-7B-Instruct~\cite{qwen2.5}. All parameters of these three modules are trainable. 

\subsection{Comparison with State-of-the-Art}
\textbf{Accuracy}: 
We compare the performance of DynImg with other methods across five open-ended video understanding datasets, as is shown in Tab.~\ref{mainexcel}. Our method demonstrates superior accuracy in most benchmarks. Specifically, DynImg surpasses the previous state-of-the-art methods by approximately 2.0\% on MSVD, MSRVTT, TGIF, and ActivityNet. 

We also assess DynImg's performance on the multi-choice question-answering dataset MVBench, whose questions emphasize comprehensive spatiotemporal understanding. As depicted in Tab.~\ref{mvbench}, our method achieves optimal performance across most evaluation problem categories, evidencing its ability to provide more comprehensible temporal information. Notably, for moving-sensitive tasks, DynImg demonstrates substantial gains in Moving Direction (+21.0\%), Moving Count (+15.0\%), and Moving Attribute (+26.5\%), directly quantifying its effectiveness for rapid object movements.

\noindent\textbf{Token Efficiency}: 
We further analyze the token-level efficiency of the proposed DynImg and existing methods, as shown on the right side of Fig.~\ref{first}. The horizontal axis reflects efficiency, which is inversely proportional to the number of visual tokens used. The vertical axis shows the average accuracy across four datasets. Our method not only achieves high accuracy but also demonstrates strong token-level efficiency. Thanks to the effective utilization of temporal information, DynImg achieves comparable accuracy using only 4 frames (DynImgs), whereas the baseline method PLLaVA requires a 16-frame input. This significantly reduces the number of visual tokens fed into the LLM, improving both training and inference efficiency without sacrificing expressive power.

While the MPEG-4 video decoding method only introduces minimal overhead compared to methods like Decord, the overall data loading time does increase from 0.06s to 0.32s. This is mainly due to the increased number of frames being read, along with operations such as resizing and composition. However, this overhead remains acceptable when compared to the much longer duration of MLLM training. Moreover, any extra memory consumed during composition is promptly released afterward.

\subsection{Ablation Studies}
\textbf{Different design choices of DynImg.}
To demonstrate the effectiveness of our proposed temporal prompts and the corresponding position embedding, we conduct ablation experiments on various design strategies of DynImg, as shown in Tab.\ref{componentablation}. In our implementation, temporal prompts are overlaid at the image level, and the composited DynImg is fed into the visual encoder for feature extraction. We compare our DynImg with an alternative, more direct method, where all keyframes and non-keyframes are input into the visual encoder, followed by similar overlay processing at the feature level. It is observed that the post-encoder prompt fusion yields limited effectiveness. We deduce the reason is that these post-encoder prompts merely add more information for LLM, without facilitating spatiotemporal interaction. In contrast, applying prompts before the encoder results in a performance improvement of about 2.4\%, highlighting the efficacy of our pre-encoder spatiotemporal interaction.

Additionally, we look into the effect of the proposed 4D RoPE. In the scenario where temporal prompts are used without the 4D positional embedding, a traditional 1D positional embedding along the sequence dimension is employed. The result indicates that for LLM, DynImg with accurate positional embedding can appropriately signal the relationships among input visual tokens, mitigating potential spatiotemporal relationship confusion caused by DynImg composition.

\begin{table}
\centering
\setlength{\tabcolsep}{8pt}
\begin{threeparttable}
\resizebox{0.4\textwidth}{!}{
\begin{tabular}{ccc |c c}
\toprule[1pt]
\multicolumn{3}{c|}{Components} &\multicolumn{2}{c}{MSVD} \\
TP&TP-Stage&4D-RoPE&Acc&Score \\
\midrule
&&&74.9&3.8   \\
\checkmark&After Encoder&&75.2&3.9   \\
\checkmark&Before Encoder&&77.3&4.0    \\
\checkmark&Before Encoder&\checkmark&\textbf{78.6}&\textbf{4.2}   \\
\bottomrule[1pt]     
\end{tabular}
}
\end{threeparttable}
\caption{Ablation studies on the key components of DynImg. ``TP'' refers to the temporal prompt. ``TP-stage'' refers to the stage at which temporal prompts are applied. In DynImg, temporal prompts are concatenated with the original image before the visual encoder. ``After Encoder'' denotes the approach where both non-keyframes and keyframes are fed into the visual encoder at the same resolution, followed by downsampling and concatenation at the feature level. ``4D-RoPE'' denotes the proposed 4D rotary position embedding. }
\label{componentablation}
\end{table}

\noindent\textbf{Different numbers of $N$-frames in one DynImg.} 
We explore the influence of the number of $N$-frames in the composition process of DynImg, as shown in Tab.~\ref{nonkeynum}. It is important to note that we perform different resizing and placing depending on the number of non-key frames. We ensure that the size of each non-key frame in the temporal prompt is an integer multiple of the patch size during feature extraction and can be represented by an integer number of tokens as the LLM input. This aims to avoid the issue of boundary spanning between key frames and non-key frames.

It can be observed that when the number of $N$-frames is too small, the effect of temporal visual prompts is poor, resulting in a lower accuracy of video understanding. This is because when the number of non-key frames is one or two, the subtle temporal variations they contain are not prominent; rather than serving as a temporal prompt, they primarily represent redundant spatial information in smaller size. When the number of $N$-frames increases to four, the temporal information starts to play an effective prompting role, the accuracy of video understanding is the highest. 
As the number of non-key frames continues to increase, the effectiveness decreases. We speculate the reason for this is as follows. In order to arrange the $N$-frames in a row, as their number increases, the size of each frame has to decrease, resulting in a reduction of information. The low resolution prevents them from effectively providing temporal prompts.

\noindent\textbf{Different numbers of DynImg.} We also investigate the influence of the DynImg number for one video, where each DynImg consists of one keyframe and four $N$-frames. As shown in Tab.~\ref{dynimgnum}, when the number of DynImgs is relatively small, the video understanding performance improves as the number of DynImgs increases. However, once the number of DynImgs exceeds four, performance either stabilizes or declines. This is because a higher number of DynImgs corresponds to a higher video sampling frequency of both $K$-frames and $N$-frames. This situation results in minimal temporal variation within frames in each DynImg's temporal prompt, which fails to provide effective temporal information supplementation for the nearly identical keyframe.

\begin{table}
    \centering
    \begin{minipage}{0.46\linewidth}
        \centering
        \resizebox{1.0\linewidth}{!}{
            \begin{tabular}{l|ccccc}
            \toprule[1pt]
            Num&1&2&4&6&12\\
            \midrule
            Acc&71.9&72.3&\textbf{78.6}&78.1&77.5\\
            Score&3.3&3.4&\textbf{4.2}&3.7&3.7\\
            \bottomrule[1pt]     
            \end{tabular}}
        \captionof{table}{Ablation studies of $N$-frame number in DynImg.}
        \label{nonkeynum}
    \end{minipage} 
    \hfill
    \begin{minipage}{0.51\linewidth}
        \centering
        \resizebox{1.0\linewidth}{!}{
            \begin{tabular}{l|cccccc}
            \toprule[1pt]
            Num&1&2&4&6&8&16\\
            \midrule
            Acc&77.0&77.8&\textbf{78.6}&78.5&78.5&77.7\\
            Score&3.9&4.1&\textbf{4.2}&4.0&3.8&3.8\\
            \bottomrule[1pt]     
            \end{tabular}}
         \caption{Ablation studies of different numbers of DynImg.}
    \label{dynimgnum}
    \end{minipage} 
\end{table}

\begin{table}[!t]
\centering
\setlength{\tabcolsep}{8pt}
\begin{threeparttable}
\resizebox{0.45\textwidth}{!}{
\begin{tabular}{l|cc}
\toprule[1pt]
Coordinate \& Operation&Acc&Score \\
\midrule
1D Coor (S)&77.3&4.0   \\
3D Coor (H,W,S) + Split ($\theta$ Fixed)&76.8&4.0   \\
3D Coor (H,W,S) + Split ($\theta$ Trainable)&77.8&4.1   \\
3D Coor (H,W,S) + Merge ($\theta$ Trainable)&78.0&4.1   \\
4D Coor (H,W,T,S) + Split ($\theta$ Fixed)&77.0&4.0   \\
4D Coor (H,W,T,S) + Split ($\theta$ Trainable)&78.4&4.2   \\
\rowcolor{gray!20}
4D Coor (H,W,T,S) + Merge ($\theta$ Trainable)&\textbf{78.6}&\textbf{4.2}   \\
\bottomrule[1pt]     
\end{tabular}
}
\end{threeparttable}
\caption{Ablation studies of different implements of position embedding. This includes the dimension of the coordinate and the operation on them. ``Split'' varies trigonometric values for each dimension. ``Merge'' means angle calculation first and addition later.``$\theta$'' is the angle embedding to multiply with coordinates, while ``Fixed'' means the coordinate of all dimensions shares the same ``$\theta$'' that exponentially decay as the formula of ``$\theta_s$''. ``Trainable'' means the $\theta_h$, $\theta_w$, (and the $\theta_t$) are trainable parameters. }
\label{ropeablation}
\end{table}

\noindent\textbf{Different implementation of position embedding.}
We also investigate the impact of different positional embedding methods. As a popular positional encoding method in LLMs, RoPE is computed by multiplying features $q$ with trigonometric functions based on coordinates $x$, as is shown in Eq.~\ref{eqation}. When the input turns from a 1D sequence to DynImg with (H,W,T,S) four dimensions, RoPE's different implementations can be explored in two aspects. 
First is the coordinate representation $x$. We can reduce the dimensions to a 1D sequence coordinate, treat DynImg as a whole image using (HWS) dimensions, or unify text and visual tokens in the 4D coordinates of DynImg.
As shown in Tab.\ref{ropeablation}, the accuracy improves with increasing coordinate dimensions under the same operation. This is because the closer the coordinate dimensions align with the actual information dimensions in DynImg, the better the LLM can interpret the data. This finding aligns with the description in Sec.3, affirming the need for the proposed 4D RoPE for DynImg.

Second is ablation of ``Operation'' to angles after obtaining the coordinates. In Tab.\ref{ropeablation}, ``Split'' means various trigonometric values for each dimension, where

\noindent{ \normalsize\( R = \begin{pmatrix} q_1 \\ q_2 \\ ... \\ q_7 \\ q_8 \end{pmatrix} \otimes \begin{pmatrix} \cos(x_h\theta_h) \\ \cos(x_h\theta_h) \\... \\\cos(x_s\theta_s) \\ \cos(x_s\theta_s) \end{pmatrix} + \begin{pmatrix} -q_2 \\ q_1 \\ ... \\ -q_8 \\ q_7\end{pmatrix} \otimes \begin{pmatrix} \sin(x_h\theta_h) \\ \sin(x_h\theta_h) \\... \\\sin(x_s\theta_s) \\ \sin(x_s\theta_s) \end{pmatrix} \)}.

\noindent``Merge'' refers to converting coordinates into angles first, and then summing them, where {\( x \cdot \theta = x_h \cdot \theta_h + x_w \cdot \theta_w + x_t \cdot \theta_t + x_s \cdot \theta_s \)}, as described in Sec3.2. In 1D RoPE, $\theta_s$ is an exponentially decaying fixed value derived from a formula. For 3D or 4D coordinates, we can let all dimensions share the same $\theta$, which implies that all RoPE parameters are non-learnable, labeled as ``Fix'' in Tab.\ref{ropeablation}. In contrast, ``Trainable'' indicates that $\theta_h,\theta_w$ (and $\theta_t$) are learnable parameters. It is found that fixed $\theta$ values lead to poor accuracy. This indicates that LLM cannot directly comprehend the positional encoding of the novel composition format of DynImg and requires training for progressive learning. For trainable $\theta$ in each dimension, both split trigonometric calculations and angle merging significantly enhance the performance, with angle merging offering slightly higher comprehension accuracy. This is because angle merging, by initializing $\theta_h,\theta_w$ (and $\theta_t$) to zero, allows $\theta_s$ to dominate during the early stages of model training. This preserves the capabilities of the pre-trained LLM in the early training phase, which benefits the model's training and inference.

\section{Conclusion}
We propose DynImg, a novel method for video representation. The integration of the proposed temporal prompts allows for an adjustment in the spatial attention mechanism, ensuring that temporally significant features are not overlooked. This adjustment mitigates the loss of fine-grained information prior to the spatio-temporal interaction. Furthermore, the 4D rotational positional embedding designed for DynImg preserves the correct spatiotemporal sequence and adjacency throughout the compositional process. Experimental results indicate that our DynImg is both effective and efficient, leading to enhanced video understanding.

{
    \small
    \bibliographystyle{ieeenat_fullname}
    \bibliography{main}

\begin{thebibliography}{57}
\providecommand{\natexlab}[1]{#1}
\providecommand{\url}[1]{\texttt{#1}}
\expandafter\ifx\csname urlstyle\endcsname\relax
  \providecommand{\doi}[1]{doi: #1}\else
  \providecommand{\doi}{doi: \begingroup \urlstyle{rm}\Url}\fi

\bibitem[Alayrac et~al.(2022)Alayrac, Donahue, Luc, Miech, Barr, Hasson, Lenc, Mensch, Millican, Reynolds, et~al.]{alayrac2022flamingo}
Jean-Baptiste Alayrac, Jeff Donahue, Pauline Luc, Antoine Miech, Iain Barr, Yana Hasson, Karel Lenc, Arthur Mensch, Katherine Millican, Malcolm Reynolds, et~al.
\newblock Flamingo: a visual language model for few-shot learning.
\newblock \emph{NeurIPS}, 2022.

\bibitem[Arnab et~al.(2021)Arnab, Dehghani, Heigold, Sun, Lu{\v{c}}i{\'c}, and Schmid]{vivit}
Anurag Arnab, Mostafa Dehghani, Georg Heigold, Chen Sun, Mario Lu{\v{c}}i{\'c}, and Cordelia Schmid.
\newblock Vivit: A video vision transformer.
\newblock In \emph{Proceedings of the IEEE/CVF international conference on computer vision}, pages 6836--6846, 2021.

\bibitem[Bahng et~al.(2022)Bahng, Jahanian, Sankaranarayanan, and Isola]{promptfirst}
Hyojin Bahng, Ali Jahanian, Swami Sankaranarayanan, and Phillip Isola.
\newblock Exploring visual prompts for adapting large-scale models.
\newblock \emph{arXiv preprint arXiv:2203.17274}, 2022.

\bibitem[Bao et~al.(2024{\natexlab{a}})Bao, Qin, Sun, Wang, and Zheng]{bao2024relevant}
Xiaoyi Bao, Jie Qin, Siyang Sun, Xingang Wang, and Yun Zheng.
\newblock Relevant intrinsic feature enhancement network for few-shot semantic segmentation.
\newblock In \emph{Proceedings of the AAAI Conference on Artificial Intelligence}, pages 765--773, 2024{\natexlab{a}}.

\bibitem[Bao et~al.(2024{\natexlab{b}})Bao, Sun, Ma, Zheng, Guo, Zhao, Zheng, and Wang]{bao2024cores}
Xiaoyi Bao, Siyang Sun, Shuailei Ma, Kecheng Zheng, Yuxin Guo, Guosheng Zhao, Yun Zheng, and Xingang Wang.
\newblock Cores: Orchestrating the dance of reasoning and segmentation.
\newblock In \emph{European Conference on Computer Vision}, pages 187--204. Springer, 2024{\natexlab{b}}.

\bibitem[Bertasius et~al.(2021)Bertasius, Wang, and Torresani]{timesformer}
Gedas Bertasius, Heng Wang, and Lorenzo Torresani.
\newblock Is space-time attention all you need for video understanding?
\newblock In \emph{ICML}, page~4, 2021.

\bibitem[Caba~Heilbron et~al.(2015)Caba~Heilbron, Escorcia, Ghanem, and Carlos~Niebles]{activitynet}
Fabian Caba~Heilbron, Victor Escorcia, Bernard Ghanem, and Juan Carlos~Niebles.
\newblock Activitynet: A large-scale video benchmark for human activity understanding.
\newblock In \emph{CVPR}, pages 961--970, 2015.

\bibitem[Carreira and Zisserman(2017)]{i3d}
Joao Carreira and Andrew Zisserman.
\newblock Quo vadis, action recognition? a new model and the kinetics dataset.
\newblock In \emph{proceedings of the IEEE Conference on Computer Vision and Pattern Recognition}, pages 6299--6308, 2017.

\bibitem[Diba et~al.(2017)Diba, Sharma, and Van~Gool]{tle}
Ali Diba, Vivek Sharma, and Luc Van~Gool.
\newblock Deep temporal linear encoding networks.
\newblock In \emph{Proceedings of the IEEE conference on Computer Vision and Pattern Recognition}, pages 2329--2338, 2017.

\bibitem[Dosovitskiy et~al.(2020)Dosovitskiy, Beyer, Kolesnikov, Weissenborn, Zhai, Unterthiner, Dehghani, Minderer, Heigold, Gelly, et~al.]{dosovitskiy2020image}
Alexey Dosovitskiy, Lucas Beyer, Alexander Kolesnikov, Dirk Weissenborn, Xiaohua Zhai, Thomas Unterthiner, Mostafa Dehghani, Matthias Minderer, Georg Heigold, Sylvain Gelly, et~al.
\newblock An image is worth 16x16 words: Transformers for image recognition at scale.
\newblock \emph{arXiv preprint arXiv:2010.11929}, 2020.

\bibitem[Feichtenhofer et~al.(2016)Feichtenhofer, Pinz, and Zisserman]{twoearly}
Christoph Feichtenhofer, Axel Pinz, and Andrew Zisserman.
\newblock Convolutional two-stream network fusion for video action recognition.
\newblock In \emph{Proceedings of the IEEE conference on computer vision and pattern recognition}, pages 1933--1941, 2016.

\bibitem[Feichtenhofer et~al.(2019)Feichtenhofer, Fan, Malik, and He]{slowfast}
Christoph Feichtenhofer, Haoqi Fan, Jitendra Malik, and Kaiming He.
\newblock Slowfast networks for video recognition.
\newblock In \emph{Proceedings of the IEEE/CVF international conference on computer vision}, pages 6202--6211, 2019.

\bibitem[Gu et~al.(2023)Gu, Han, Chen, Beirami, He, Zhang, Liao, Qin, Tresp, and Torr]{prompt8}
Jindong Gu, Zhen Han, Shuo Chen, Ahmad Beirami, Bailan He, Gengyuan Zhang, Ruotong Liao, Yao Qin, Volker Tresp, and Philip Torr.
\newblock A systematic survey of prompt engineering on vision-language foundation models.
\newblock \emph{arXiv preprint arXiv:2307.12980}, 2023.

\bibitem[Huang et~al.(2024)Huang, Liao, Radhakrishnan, Yin, Molchanov, Yu, and Kautz]{huang2024lita}
De-An Huang, Shijia Liao, Subhashree Radhakrishnan, Hongxu Yin, Pavlo Molchanov, Zhiding Yu, and Jan Kautz.
\newblock Lita: Language instructed temporal-localization assistant.
\newblock In \emph{European Conference on Computer Vision}, pages 202--218. Springer, 2024.

\bibitem[Jia et~al.(2022)Jia, Tang, Chen, Cardie, Belongie, Hariharan, and Lim]{prompt11}
Menglin Jia, Luming Tang, Bor-Chun Chen, Claire Cardie, Serge Belongie, Bharath Hariharan, and Ser-Nam Lim.
\newblock Visual prompt tuning.
\newblock In \emph{European Conference on Computer Vision}, pages 709--727. Springer, 2022.

\bibitem[Jin et~al.(2024{\natexlab{a}})Jin, Takanobu, Zhang, Cao, and Yuan]{jin2024chatuniv}
Peng Jin, Ryuichi Takanobu, Wancai Zhang, Xiaochun Cao, and Li Yuan.
\newblock Chat-univi: Unified visual representation empowers large language models with image and video understanding.
\newblock In \emph{Proceedings of the IEEE/CVF Conference on Computer Vision and Pattern Recognition}, pages 13700--13710, 2024{\natexlab{a}}.

\bibitem[Jin et~al.(2024{\natexlab{b}})Jin, Sun, Xu, Chen, Jiang, Huang, Song, Liu, Zhang, Song, et~al.]{jin2024videolavit}
Yang Jin, Zhicheng Sun, Kun Xu, Liwei Chen, Hao Jiang, Quzhe Huang, Chengru Song, Yuliang Liu, Di Zhang, Yang Song, et~al.
\newblock Video-lavit: Unified video-language pre-training with decoupled visual-motional tokenization.
\newblock \emph{arXiv preprint arXiv:2402.03161}, 2024{\natexlab{b}}.

\bibitem[Karpathy et~al.(2014)Karpathy, Toderici, Shetty, Leung, Sukthankar, and Fei-Fei]{deepvideo}
Andrej Karpathy, George Toderici, Sanketh Shetty, Thomas Leung, Rahul Sukthankar, and Li Fei-Fei.
\newblock Large-scale video classification with convolutional neural networks.
\newblock In \emph{Proceedings of the IEEE conference on Computer Vision and Pattern Recognition}, pages 1725--1732, 2014.

\bibitem[Khattak et~al.(2023)Khattak, Rasheed, Maaz, Khan, and Khan]{khattak2023maple}
Muhammad~Uzair Khattak, Hanoona Rasheed, Muhammad Maaz, Salman Khan, and Fahad~Shahbaz Khan.
\newblock Maple: Multi-modal prompt learning.
\newblock In \emph{Proceedings of the IEEE/CVF Conference on Computer Vision and Pattern Recognition}, pages 19113--19122, 2023.

\bibitem[Kim et~al.(2024)Kim, Choi, Lee, and Rhee]{kim2024IGVLM}
Wonkyun Kim, Changin Choi, Wonseok Lee, and Wonjong Rhee.
\newblock An image grid can be worth a video: Zero-shot video question answering using a vlm.
\newblock \emph{arXiv preprint arXiv:2403.18406}, 2024.

\bibitem[Lai et~al.(2023)Lai, Tian, Chen, Li, Yuan, Liu, and Jia]{lai2023lisa}
Xin Lai, Zhuotao Tian, Yukang Chen, Yanwei Li, Yuhui Yuan, Shu Liu, and Jiaya Jia.
\newblock Lisa: Reasoning segmentation via large language model.
\newblock \emph{arXiv preprint arXiv:2308.00692}, 2023.

\bibitem[Lan et~al.(2017)Lan, Zhu, Hauptmann, and Newsam]{dovf}
Zhenzhong Lan, Yi Zhu, Alexander~G Hauptmann, and Shawn Newsam.
\newblock Deep local video feature for action recognition.
\newblock In \emph{Proceedings of the IEEE conference on computer vision and pattern recognition workshops}, pages 1--7, 2017.

\bibitem[Li et~al.(2023{\natexlab{a}})Li, Li, Savarese, and Hoi]{li2023blip}
Junnan Li, Dongxu Li, Silvio Savarese, and Steven Hoi.
\newblock Blip-2: Bootstrapping language-image pre-training with frozen image encoders and large language models.
\newblock \emph{arXiv:2301.12597}, 2023{\natexlab{a}}.

\bibitem[Li et~al.(2023{\natexlab{b}})Li, He, Wang, Li, Wang, Luo, Wang, Wang, and Qiao]{li2023videochat}
KunChang Li, Yinan He, Yi Wang, Yizhuo Li, Wenhai Wang, Ping Luo, Yali Wang, Limin Wang, and Yu Qiao.
\newblock Videochat: Chat-centric video understanding.
\newblock \emph{arXiv preprint arXiv:2305.06355}, 2023{\natexlab{b}}.

\bibitem[Li et~al.(2024)Li, Wang, He, Li, Wang, Liu, Wang, Xu, Chen, Luo, et~al.]{li2024mvbench}
Kunchang Li, Yali Wang, Yinan He, Yizhuo Li, Yi Wang, Yi Liu, Zun Wang, Jilan Xu, Guo Chen, Ping Luo, et~al.
\newblock Mvbench: A comprehensive multi-modal video understanding benchmark.
\newblock In \emph{Proceedings of the IEEE/CVF Conference on Computer Vision and Pattern Recognition}, pages 22195--22206, 2024.

\bibitem[Li et~al.(2016)Li, Song, Cao, Tetreault, Goldberg, Jaimes, and Luo]{li2016tgif}
Yuncheng Li, Yale Song, Liangliang Cao, Joel Tetreault, Larry Goldberg, Alejandro Jaimes, and Jiebo Luo.
\newblock Tgif: A new dataset and benchmark on animated gif description.
\newblock In \emph{Proceedings of the IEEE Conference on Computer Vision and Pattern Recognition}, pages 4641--4650, 2016.

\bibitem[Li et~al.(2025)Li, Wang, and Jia]{li2025llamavid}
Yanwei Li, Chengyao Wang, and Jiaya Jia.
\newblock Llama-vid: An image is worth 2 tokens in large language models.
\newblock In \emph{European Conference on Computer Vision}, pages 323--340. Springer, 2025.

\bibitem[Lin et~al.(2023)Lin, Ye, Zhu, Cui, Ning, Jin, and Yuan]{lin2023videollava}
Bin Lin, Yang Ye, Bin Zhu, Jiaxi Cui, Munan Ning, Peng Jin, and Li Yuan.
\newblock Video-llava: Learning united visual representation by alignment before projection.
\newblock \emph{arXiv preprint arXiv:2311.10122}, 2023.

\bibitem[Liu et~al.(2024{\natexlab{a}})Liu, Li, Wu, and Lee]{llava}
Haotian Liu, Chunyuan Li, Qingyang Wu, and Yong~Jae Lee.
\newblock Visual instruction tuning.
\newblock \emph{Advances in neural information processing systems}, 36, 2024{\natexlab{a}}.

\bibitem[Liu et~al.(2023)Liu, Yuan, Fu, Jiang, Hayashi, and Neubig]{promptnlp}
Pengfei Liu, Weizhe Yuan, Jinlan Fu, Zhengbao Jiang, Hiroaki Hayashi, and Graham Neubig.
\newblock Pre-train, prompt, and predict: A systematic survey of prompting methods in natural language processing.
\newblock \emph{ACM Computing Surveys}, 55\penalty0 (9):\penalty0 1--35, 2023.

\bibitem[Liu et~al.(2024{\natexlab{b}})Liu, Li, Tang, Ge, Shan, and Li]{liu2024stllm}
Ruyang Liu, Chen Li, Haoran Tang, Yixiao Ge, Ying Shan, and Ge Li.
\newblock St-llm: Large language models are effective temporal learners.
\newblock In \emph{European Conference on Computer Vision}, pages 1--18. Springer, 2024{\natexlab{b}}.

\bibitem[Liu et~al.(2024{\natexlab{c}})Liu, Tang, Liu, Ge, Shan, Li, and Yang]{liu2024ppllava}
Ruyang Liu, Haoran Tang, Haibo Liu, Yixiao Ge, Ying Shan, Chen Li, and Jiankun Yang.
\newblock Ppllava: Varied video sequence understanding with prompt guidance.
\newblock \emph{arXiv preprint arXiv:2411.02327}, 2024{\natexlab{c}}.

\bibitem[Liu et~al.(2025)Liu, Xie, Li, Zhao, Tang, Zheng, Liu, and Xie]{liu2025hybrid}
Zhihang Liu, Chen-Wei Xie, Pandeng Li, Liming Zhao, Longxiang Tang, Yun Zheng, Chuanbin Liu, and Hongtao Xie.
\newblock Hybrid-level instruction injection for video token compression in multi-modal large language models.
\newblock In \emph{Proceedings of the Computer Vision and Pattern Recognition Conference}, pages 8568--8578, 2025.

\bibitem[Ma et~al.(2023)Ma, Jin, Wang, Xian, Feng, and Yang]{ma2023vista}
Fan Ma, Xiaojie Jin, Heng Wang, Yuchen Xian, Jiashi Feng, and Yi Yang.
\newblock Vista-llama: Reliable video narrator via equal distance to visual tokens.
\newblock \emph{arXiv preprint arXiv:2312.08870}, 2023.

\bibitem[Maaz et~al.(2024)Maaz, Rasheed, Khan, and Khan]{maaz2023videochatgpt}
Muhammad Maaz, Hanoona Rasheed, Salman Khan, and Fahad~Shahbaz Khan.
\newblock Video-chatgpt: Towards detailed video understanding via large vision and language models.
\newblock In \emph{Proceedings of the 62nd Annual Meeting of the Association for Computational Linguistics (ACL 2024)}, 2024.

\bibitem[Radford et~al.(2021)Radford, Kim, Hallacy, Ramesh, Goh, Agarwal, Sastry, Askell, Mishkin, Clark, et~al.]{clip}
Alec Radford, Jong~Wook Kim, Chris Hallacy, Aditya Ramesh, Gabriel Goh, Sandhini Agarwal, Girish Sastry, Amanda Askell, Pamela Mishkin, Jack Clark, et~al.
\newblock Learning transferable visual models from natural language supervision.
\newblock In \emph{International conference on machine learning}, pages 8748--8763. PmLR, 2021.

\bibitem[Shen et~al.(2024)Shen, Yang, Zhang, Zhai, Gonzalez, Keutzer, and Darrell]{prompt31}
Sheng Shen, Shijia Yang, Tianjun Zhang, Bohan Zhai, Joseph~E Gonzalez, Kurt Keutzer, and Trevor Darrell.
\newblock Multitask vision-language prompt tuning.
\newblock In \emph{Proceedings of the IEEE/CVF Winter Conference on Applications of Computer Vision}, pages 5656--5667, 2024.

\bibitem[Shtedritski et~al.(2023)Shtedritski, Rupprecht, and Vedaldi]{prompt32}
Aleksandar Shtedritski, Christian Rupprecht, and Andrea Vedaldi.
\newblock What does clip know about a red circle? visual prompt engineering for vlms.
\newblock In \emph{Proceedings of the IEEE/CVF International Conference on Computer Vision}, pages 11987--11997, 2023.

\bibitem[Simonyan and Zisserman(2014)]{twostream}
Karen Simonyan and Andrew Zisserman.
\newblock Two-stream convolutional networks for action recognition in videos.
\newblock \emph{Advances in neural information processing systems}, 27, 2014.

\bibitem[Song et~al.(2024)Song, Chai, Wang, Zhang, Zhou, Wu, Chi, Guo, Ye, Zhang, et~al.]{song2024moviechat}
Enxin Song, Wenhao Chai, Guanhong Wang, Yucheng Zhang, Haoyang Zhou, Feiyang Wu, Haozhe Chi, Xun Guo, Tian Ye, Yanting Zhang, et~al.
\newblock Moviechat: From dense token to sparse memory for long video understanding.
\newblock In \emph{Proceedings of the IEEE/CVF Conference on Computer Vision and Pattern Recognition}, pages 18221--18232, 2024.

\bibitem[Tang et~al.(2025)Tang, Xie, Wang, Bao, Weng, Li, Zheng, and Wang]{tang2025ufo}
Hao Tang, Chenwei Xie, Haiyang Wang, Xiaoyi Bao, Tingyu Weng, Pandeng Li, Yun Zheng, and Liwei Wang.
\newblock Ufo: A unified approach to fine-grained visual perception via open-ended language interface.
\newblock \emph{arXiv preprint arXiv:2503.01342}, 2025.

\bibitem[Team(2024)]{qwen2.5}
Qwen Team.
\newblock Qwen2.5: A party of foundation models, 2024.

\bibitem[Tran et~al.(2015)Tran, Bourdev, Fergus, Torresani, and Paluri]{c3d}
Du Tran, Lubomir Bourdev, Rob Fergus, Lorenzo Torresani, and Manohar Paluri.
\newblock Learning spatiotemporal features with 3d convolutional networks.
\newblock In \emph{Proceedings of the IEEE international conference on computer vision}, pages 4489--4497, 2015.

\bibitem[Tran et~al.(2018)Tran, Wang, Torresani, Ray, LeCun, and Paluri]{rtwooned}
Du Tran, Heng Wang, Lorenzo Torresani, Jamie Ray, Yann LeCun, and Manohar Paluri.
\newblock A closer look at spatiotemporal convolutions for action recognition.
\newblock In \emph{Proceedings of the IEEE conference on Computer Vision and Pattern Recognition}, pages 6450--6459, 2018.

\bibitem[Wang et~al.(2015)Wang, Qiao, and Tang]{tdd}
Limin Wang, Yu Qiao, and Xiaoou Tang.
\newblock Action recognition with trajectory-pooled deep-convolutional descriptors.
\newblock In \emph{Proceedings of the IEEE conference on computer vision and pattern recognition}, pages 4305--4314, 2015.

\bibitem[Wang et~al.(2016)Wang, Xiong, Wang, Qiao, Lin, Tang, and Van~Gool]{tsn}
Limin Wang, Yuanjun Xiong, Zhe Wang, Yu Qiao, Dahua Lin, Xiaoou Tang, and Luc Van~Gool.
\newblock Temporal segment networks: Towards good practices for deep action recognition.
\newblock In \emph{European conference on computer vision}, pages 20--36. Springer, 2016.

\bibitem[Wu et~al.(2017)Wu, Yao, Fu, and Jiang]{msvd}
Zuxuan Wu, Ting Yao, Yanwei Fu, and Yu-Gang Jiang.
\newblock Deep learning for video classification and captioning.
\newblock In \emph{Frontiers of multimedia research}, pages 3--29. 2017.

\bibitem[Xu et~al.(2016)Xu, Mei, Yao, and Rui]{msrvtt}
Jun Xu, Tao Mei, Ting Yao, and Yong Rui.
\newblock Msr-vtt: A large video description dataset for bridging video and language.
\newblock In \emph{CVPR}, pages 5288--5296, 2016.

\bibitem[Xu et~al.(2024)Xu, Zhao, Zhou, Lin, Ng, and Feng]{xu2024pllava}
Lin Xu, Yilin Zhao, Daquan Zhou, Zhijie Lin, See~Kiong Ng, and Jiashi Feng.
\newblock Pllava: Parameter-free llava extension from images to videos for video dense captioning.
\newblock \emph{arXiv preprint arXiv:2404.16994}, 2024.

\bibitem[Yue-Hei~Ng et~al.(2015)Yue-Hei~Ng, Hausknecht, Vijayanarasimhan, Vinyals, Monga, and Toderici]{twolstm}
Joe Yue-Hei~Ng, Matthew Hausknecht, Sudheendra Vijayanarasimhan, Oriol Vinyals, Rajat Monga, and George Toderici.
\newblock Beyond short snippets: Deep networks for video classification.
\newblock In \emph{Proceedings of the IEEE conference on computer vision and pattern recognition}, pages 4694--4702, 2015.

\bibitem[Zang et~al.(2022)Zang, Li, Zhou, Huang, and Loy]{prompt44}
Yuhang Zang, Wei Li, Kaiyang Zhou, Chen Huang, and Chen~Change Loy.
\newblock Unified vision and language prompt learning.
\newblock \emph{arXiv preprint arXiv:2210.07225}, 2022.

\bibitem[Zhai et~al.(2023)Zhai, Mustafa, Kolesnikov, and Beyer]{siglip}
Xiaohua Zhai, Basil Mustafa, Alexander Kolesnikov, and Lucas Beyer.
\newblock Sigmoid loss for language image pre-training.
\newblock In \emph{ICCV}, pages 11975--11986, 2023.

\bibitem[Zhang et~al.(2016)Zhang, Wang, Wang, Qiao, and Wang]{motion1}
Bowen Zhang, Limin Wang, Zhe Wang, Yu Qiao, and Hanli Wang.
\newblock Real-time action recognition with enhanced motion vector cnns.
\newblock In \emph{Proceedings of the IEEE conference on computer vision and pattern recognition}, pages 2718--2726, 2016.

\bibitem[Zhang et~al.(2023)Zhang, Li, and Bing]{zhang2023videollama}
Hang Zhang, Xin Li, and Lidong Bing.
\newblock Video-llama: An instruction-tuned audio-visual language model for video understanding.
\newblock \emph{arXiv preprint arXiv:2306.02858}, 2023.

\bibitem[Zhou et~al.(2022{\natexlab{a}})Zhou, Yang, Loy, and Liu]{prompt42}
Kaiyang Zhou, Jingkang Yang, Chen~Change Loy, and Ziwei Liu.
\newblock Conditional prompt learning for vision-language models.
\newblock In \emph{Proceedings of the IEEE/CVF conference on computer vision and pattern recognition}, pages 16816--16825, 2022{\natexlab{a}}.

\bibitem[Zhou et~al.(2022{\natexlab{b}})Zhou, Yang, Loy, and Liu]{prompt45}
Kaiyang Zhou, Jingkang Yang, Chen~Change Loy, and Ziwei Liu.
\newblock Learning to prompt for vision-language models.
\newblock \emph{International Journal of Computer Vision}, 130\penalty0 (9):\penalty0 2337--2348, 2022{\natexlab{b}}.

\bibitem[Zhu et~al.(2023)Zhu, Chen, Shen, Li, and Elhoseiny]{zhu2023minigpt}
Deyao Zhu, Jun Chen, Xiaoqian Shen, Xiang Li, and Mohamed Elhoseiny.
\newblock Minigpt-4: Enhancing vision-language understanding with advanced large language models.
\newblock \emph{arXiv:2304.10592}, 2023.

\end{thebibliography}
}


\end{document}